\documentclass{article}
\usepackage{ijcai24}

\usepackage{times}
\usepackage{soul}
\usepackage{url}
\usepackage[hidelinks,urlcolor=blue,linkcolor=blue,colorlinks=true,citecolor=black]{hyperref}
\usepackage[utf8]{inputenc}
\usepackage[small]{caption}
\usepackage{graphicx}
\usepackage{amsmath}
\usepackage{amsthm}
\usepackage{amssymb}%
\usepackage{booktabs}
\usepackage{algpseudocode}
\usepackage[ruled,lined,noend,algo2e]{algorithm2e}
\usepackage{algorithm}
\usepackage[switch]{lineno}
\usepackage{listings}
\usepackage{pifont}%
\usepackage{subcaption}
\usepackage{xspace}
\usepackage{bbm}
\usepackage{mathtools,xparse}

\DeclarePairedDelimiter{\norm}{\lVert}{\rVert}
\NewDocumentCommand{\normL}{ s O{} m }{%
  \IfBooleanTF{#1}{\norm*{#3}}{\norm[#2]{#3}}_{L_2(\Omega)}%
}

\newcommand{\cmark}{\ding{51}}%
\newcommand{\xmark}{\ding{55}}%

\SetKwInput{kwInput}{Input}
\SetKwInput{kwOutput}{Output}
\SetKwInput{kwInitialize}{Initialize}
\SetKwInput{kwReInitialize}{ReInitialize}
\SetKwInput{kwFreeze}{Freeze}
\SetKwBlock{StageOne}{Stage-1}{End}
\SetKwBlock{StageTwo}{Stage-2}{End}

\title{Are They the Same Picture? Adapting Concept Bottleneck Models for Human-AI Collaboration in Image Retrieval}%

\author{
Vaibhav Balloli$^1$
\and
Sara Beery$^2$\and
Elizabeth Bondi-Kelly$^1$
\affiliations
$^1$University of Michigan\\
$^2$Massachusetts Institute of Technology\\
\emails
vballoli@umich.edu,
beery@mit.edu,
ecbk@umich.edu
}

\newcommand{\pname}{\texttt{CHAIR}\xspace}
\newcommand{\methodname}{\texttt{CHAIR}\xspace}

\begin{document}

\maketitle

\begin{abstract}
    Image retrieval plays a pivotal role in applications from wildlife conservation to healthcare, for finding individual animals or relevant images to aid diagnosis. Although deep learning techniques for image retrieval have advanced significantly, their imperfect real-world performance often necessitates including human expertise. Human-in-the-loop approaches typically rely on humans completing the task independently and then combining their opinions with an AI model in various ways, as these models offer very little interpretability or \textit{correctability}. To allow humans to intervene in the AI model instead, thereby saving human time and effort, we adapt the Concept Bottleneck Model (CBM) and propose \texttt{CHAIR}. \texttt{CHAIR} (a) enables humans to correct intermediate concepts, which helps \textit{improve}  embeddings generated, and (b) allows for flexible levels of intervention that accommodate varying levels of human expertise for better retrieval. To show the efficacy of \texttt{CHAIR}, we demonstrate that our method performs better than similar models on image retrieval metrics without any external intervention. Furthermore, we also showcase how human intervention helps further improve retrieval performance, thereby achieving human-AI complementarity. 
\end{abstract}

\section{Introduction}
\label{sec:intro}

Recent advances in AI have shown great promise in important, often high-risk domains like healthcare \cite{peiffer2020machine,rajpurkar2020chexaid}, %
wildlife conservation \cite{kulits2021elephantbook,beery2019efficient}, and reducing misinformation \cite{mendes2022human}. %
However, these advances are imperfect, %
and can lead to harm when deployed. For example, \cite{beede2020human} reports errors when deploying AI models to detect diabetic retinopathy due to challenging real-world factors like lighting, leading to potential human harms. %

Researchers have proposed human-AI collaboration as a promising approach to mitigate the shortcomings of AI models in these domains \cite{de2020case}. For example, prior work in health AI has sought to achieve %
better accuracy %
via decision-support tools for clinicians \cite{peiffer2020machine} that \textit{assist} humans in decision-making. Human-in-the-loop participatory systems have also been proposed to accurately and robustly categorize wildlife images \cite{kulits2021elephantbook,miao2021iterative,bondi2022role} and to aid fact-checkers \cite{nguyen2018believe,mendes2022human}. Concept Bottleneck Models (CBMs) are another recent and promising method to facilitate collaboration \cite{koh2020concept}. They allow humans to \textit{interact} with AI models by viewing and manipulating intermediate, high-level concepts (e.g., whether a bird has a blue wing), which are then used for prediction. %

\begin{figure*}[ht!]
    \begin{subfigure}{0.49\textwidth}
        \includegraphics[width=\linewidth]{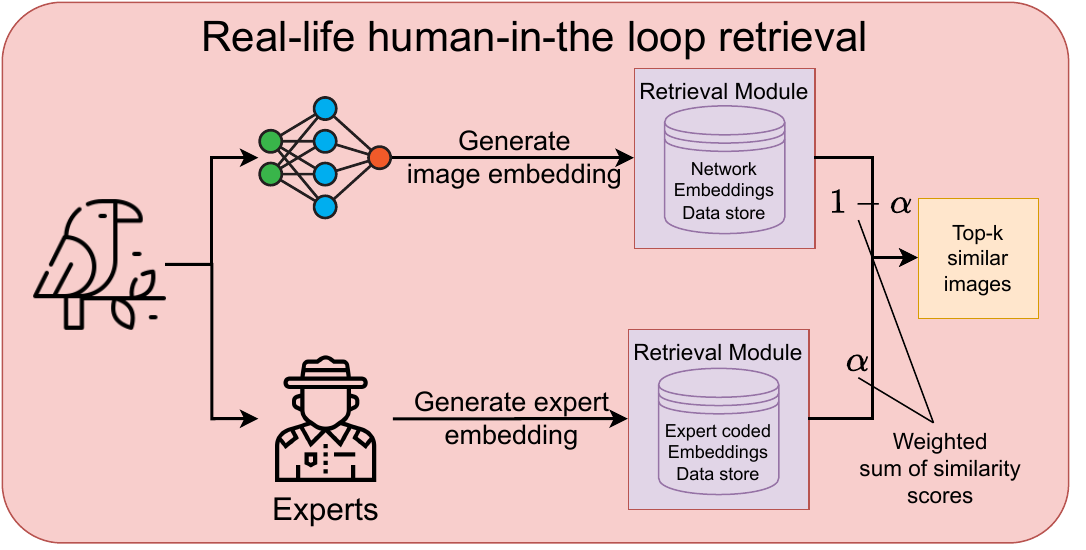} 
        \caption{Typical Human-in-the-loop pipeline for image retrieval.}
        \label{fig:hitl-retrieval}
    \end{subfigure}
    ~
    \begin{subfigure}{0.49\textwidth}
        \includegraphics[width=\linewidth]{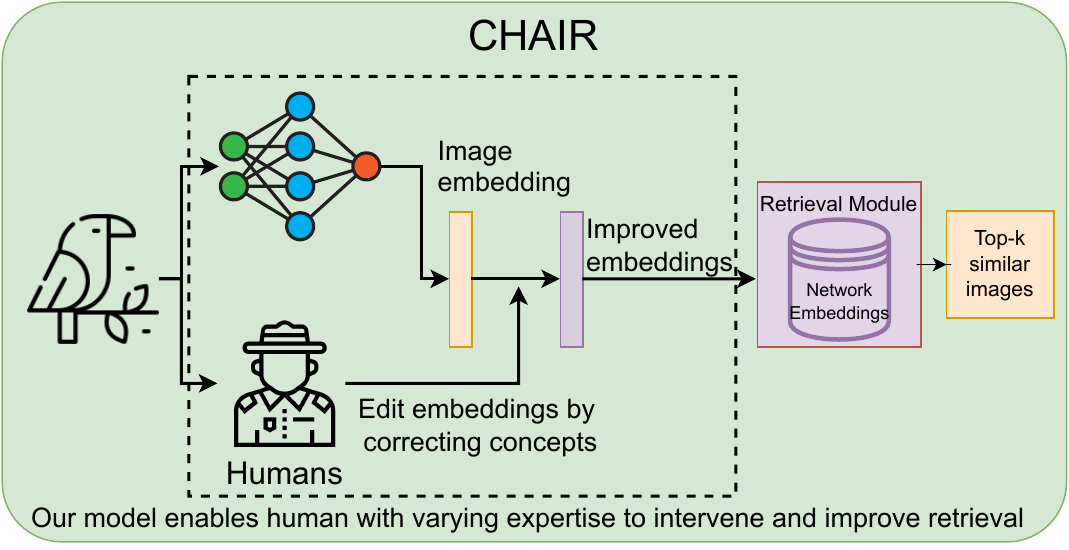}
        \caption{Our proposed collaborative pipeline for image retrieval.}
        \label{fig:hitl-retrieval-ours}
    \end{subfigure}
    
    \caption{Our proposed method allows human-AI collaboration in image retrieval by enabling humans to edit embeddings by correcting them through high-level concepts. We also enable flexible intervention to lower the expertise needed to participate.}
    \label{fig:hitl-retrieval-comparison}
    
\end{figure*}

While recent works using CBMs have significantly improved performance in classification \cite{zarlenga2023learning}, %
we note that there exist many non-classification application areas that also require human-AI collaboration. %
For example, ElephantBook \cite{kulits2021elephantbook} is a state-of-the-art elephant re-identification platform -- meant to determine which individual elephant is depicted in each image -- that adopts a semi-automated human-in-the-loop approach. This approach retrieves and presents to a user the most likely matches to individuals in a known population database for each new elephant sighting. They combine a neural network-based visual similarity ranking with a similarity ranking based on %
domain expert input via a weighted average (see Figure \ref{fig:hitl-retrieval}). However, the weighted average approach requires hyperparameter tuning ($\alpha$) depending on each user's expertise. %
Enabling humans to leverage concepts that they already use for retrieval to instead interact with the neural network in the embedding space %
has the potential to significantly reduce the amount of tuning required for each user while still including human inputs to improve team performance, as illustrated in Figure \ref{fig:hitl-retrieval-ours}). 

In this work, we aim to adapt CBMs to facilitate close human-AI collaboration at deployment time for image retrieval tasks, such as  %
ElephantBook. %
We specifically aim to answer the following research questions: \textbf{(RQ1)} How do representations generated by CBMs compare to corresponding traditional models? \textbf{(RQ2)} How can we augment CBMs to enable human intervention in image retrieval and classification? \textbf{(RQ3)} How can we train these models to incorporate varying levels of expertise? To answer these questions, we analyze how CBMs compare with traditional models for image retrieval, introduce \methodname \texttt{(CBM-Enabled Human-AI Collaboration for Image Retrieval)}, a novel CBM architecture that allows humans to intervene and encode concepts to improve retrieval, and perform extensive evaluations on how these interventions enable \textit{embedding-level} human-AI collaboration.%

\section{Background and Motivation}
\label{sec:background}

\paragraph{Concept Bottleneck Models:} The key promise of CBMs is two-fold: CBMs a) predict  \textit{high-level} intermediate concepts, which are then used to predict the final class label, thus improving interpretability, and b) enable humans to intervene and correct these intermediate concepts to improve classification performance, thus providing intervenability. This is achieved in two steps: i) \texttt{Concept Bottleneck}, which predicts the high-level, understandable concepts, and ii) \texttt{Classifier}, which predicts the final class based on the predicted concepts (see Figure \ref{fig:CBMExtend}). Human-AI collaboration is made possible here by allowing humans to \textit{correct} the model by modifying these intermediate concepts. This is shown to increase the overall classification performance on various tasks \cite{koh2020concept}.
\paragraph{Image retrieval:} Image retrieval is a key component in visual tasks like image re-identification \cite{wang2020re}, wildlife conservation \cite{kulits2021elephantbook}, remote sensing \cite{liu2020similarity}, and visual recommendation systems \cite{shankar2017deep}. Image retrieval requires systems to \textit{fetch} the most relevant images from a database given a query image, where relevance is defined depending on the application. Traditionally, applications using deep learning techniques to perform image retrieval leverage (latent) \textit{embeddings} generated by neural network trained for classification. For a given query image, a \textit{query embedding} is generated using the same neural network to find the top-$k$ nearest embeddings using some distance function \cite{wan2014deep}, typically cosine distance. We refer to the images and embeddings that are searched over as \textit{gallery} images and embeddings, respectively. Following previous literature, we utilize the \texttt{Recall@k} metric and \texttt{RecallAccuracy@k} metrics to measure the performance of an image retrieval technique. For a given image and label pair $(x_i,y_i)$, let the \texttt{top-k} retrieved images be $\mathbf{y'}_i \in \mathbf{R}^k$. Then, the \texttt{Recall@k} and \texttt{RecallAccuracy@k} are defined as follows:
\begin{align}
\begin{split}
    \text{\texttt{Recall@k}} = \frac{\sum_{i=1}^{N} \norm{y_i == \mathbf{y'}_i}_\infty}{N} 
    \label{eq:recallk}
\end{split} \\
\begin{split}
    \text{\texttt{RecallAccuracy@k}} = \frac{\sum_{i=1}^{N} \norm{y_i == \mathbf{y'}_i}_0}{N*k} 
    \label{eq:recallaccuracyk}
\end{split}
\end{align}

The \texttt{Recall@k} metric evaluates if there exists at least one \textit{accurate} image in the \texttt{top-k} images that were retrieved, whereas the \texttt{RecallAccuracy@k} evaluates the number of accurate images retrieved in the top-$K$ images. 

\paragraph{Motivation of our work:} Current methods in image retrieval provide little room for human-AI collaboration. Platforms like ElephantBook, a practical, state-of-the-art computer vision system, still require a human-in-the-loop to achieve reliable performance, thus highlighting the importance of human-AI collaboration. Our contributions in this paper leverage CBMs to enable human-AI collaboration in image retrieval tasks. Previous research in CBMs until now focus largely on improving performance by proposing different architectures \cite{EspinosaZarlenga2022cem,kim2023probabilistic,marconato2022glancenets}, modify loss functions \cite{sheth2024auxiliary}, mitigate leakage \cite{havasi2022addressing}, improve intervenability \cite{zarlenga2023learning,marcinkevivcs2024beyond} or circumvent requirements of labels \cite{oikarinen2023label}. In this work, our novelty lies in expanding \textit{intervenability} capabilities of CBMs to tasks like image retrieval, which require capturing \textit{corrected} concepts into the embedding. All the previously mentioned works are orthogonal to the aim of this paper and are complementary in improving the performance of CBMs.

\section{Do CBMs Already Work for Image Retrieval?} 
\begin{figure}%
    \centering
    \includegraphics[width=\linewidth]{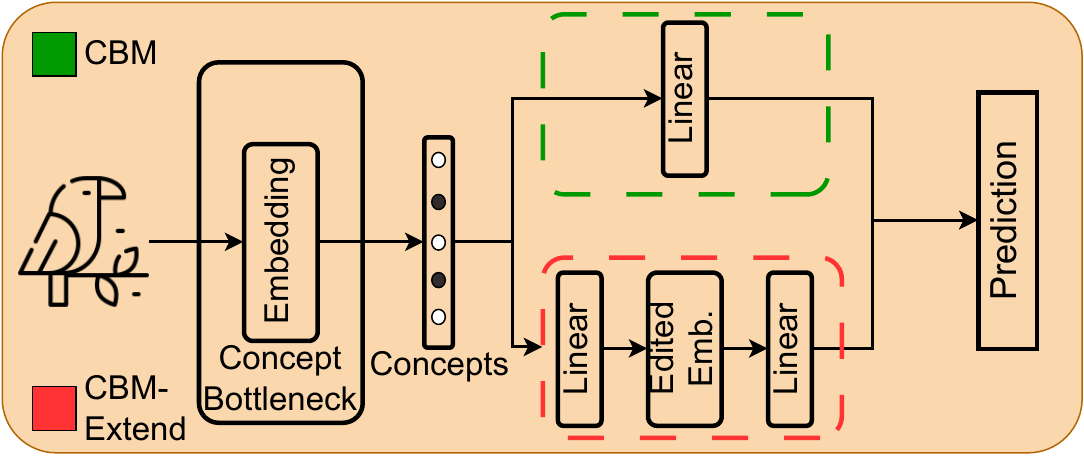}
    \caption{Illustration of the CBM and CBM-Extend, a naive extension of CBM to correct embeddings for retrieval (Edited here refers to capturing human intervention)}
    \label{fig:CBMExtend}
\end{figure}

This section aims to establish the difficulties faced in adopting single-agent (only AI or human) approaches and show how human-AI teams, specifically through CBMs, can potentially resolve some of these issues. \\

\paragraph{Neural networks - irremediable by humans:} The key drawback of using neural networks is that when these models make mistakes or inaccurate inferences, their architecture and design allow little room for intermediate human oversight, especially for tasks like image retrieval where the inaccuracies are not obvious until the results are inspected. %

\paragraph{Expert coding - costly and high barrier of entry: } Hand-crafted features and coding of retrieval systems require advanced domain knowledge and experience, which raises the barrier of entry for inexperienced users to utilize such systems. Furthermore, these codes often require entire feature sets to achieve perfect retrieval. Any ambiguity in identifying these codes, for example, due to ambiguity in or occlusion of concept-relevant parts of the image, is marked as \textit{Wildcards} in such systems. %
Since these codes typically lie on a spectrum of easily identifiable to years of expertise needed, we can leverage neural networks to alleviate some of these difficulties while having a variable level of human oversight and help reduce wildcard entries with a collaborative human-AI retrieval system. 
\paragraph{Are CBMs the solution? Potentially: }%
Our contributions stem from the observation that high-level concepts are analogous to the codes developed by domain experts to help humans retrieve relevant images. However, adopting these concepts (in the case of CBMs) or expert codes (in the case of humans) to enable collaborative embedding generation for image retrieval is not supported by existing CBM architectures (see Table \ref{tab:compare}). %

\begin{table}[ht!]
    \centering
    \resizebox{\columnwidth}{!}{%
    \begin{tabular}{|c|c|c|c|}
    \hline
    \textbf{Feature/Model} & Standard & CBMs & Ideal \\
    \hline
    Usable for retrieval & \cmark & \cmark & \cmark \\
    \hline
    Interpretable & \xmark & \cmark & \cmark \\
    \hline
    Intervenable for classification & \xmark & \cmark & \cmark \\
    \hline
    Intervenable for retrieval & \xmark & \xmark & \cmark \\
    \hline
    \end{tabular}
    }
    \caption{Ideal Model Features}
    \label{tab:compare}
\end{table}

\begin{figure}%
\centering
\includegraphics[width=0.9\linewidth]{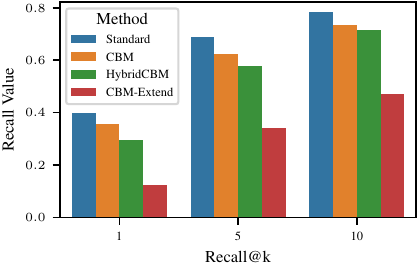}
\caption{CBM \protect\cite{koh2020concept}, HybridCBM \protect\cite{mahinpei2021promises} and Naive CBM extension (Figure~\ref{fig:CBMExtend}) have poor retrieval performance when compared to their standard counterpart model}
\label{fig:naive}
\end{figure}

CBMs typically contain a linear layer as a classifier, and the predictions change as a function of predicted concepts and human interventions. Therefore, a straightforward way of achieving intervenability for retrieval would be to extract the latent embeddings for comparison with other images after the concept predictions by adding another linear layer before predicting the final class. This enables capturing human intervention in the embedding space. However, this extension, termed as \texttt{CBM-Extend} (illustrated in Figure \ref{fig:CBMExtend}), performs poorly in image retrieval when compared to the latent embeddings used from standard, CBM and HybridCBM \cite{mahinpei2021promises} embeddings with similar base architectures (ResNet-18). (see Figure \ref{fig:naive}). The significant drop in performance (Recall@k here, as defined in equation \ref{eq:recallk}) is likely due to the lack of generalization with an increasing number of hidden layers (increased capacity), rendering the naive extensions unusable. While section \ref{sec:results} details what \texttt{Recall@k} signifies here, it is sufficient to say that there exists a clear gap in abilities and performance, thus giving a clear picture of how representations generated by CBMs compare to corresponding traditional models (\textbf{(RQ1)}). %

\section{Our Proposed Architecture: CHAIR}
\label{sec:fusion}

\begin{figure*}[ht!]
    \centering
    \includegraphics[width=0.9\linewidth,scale=0.9]{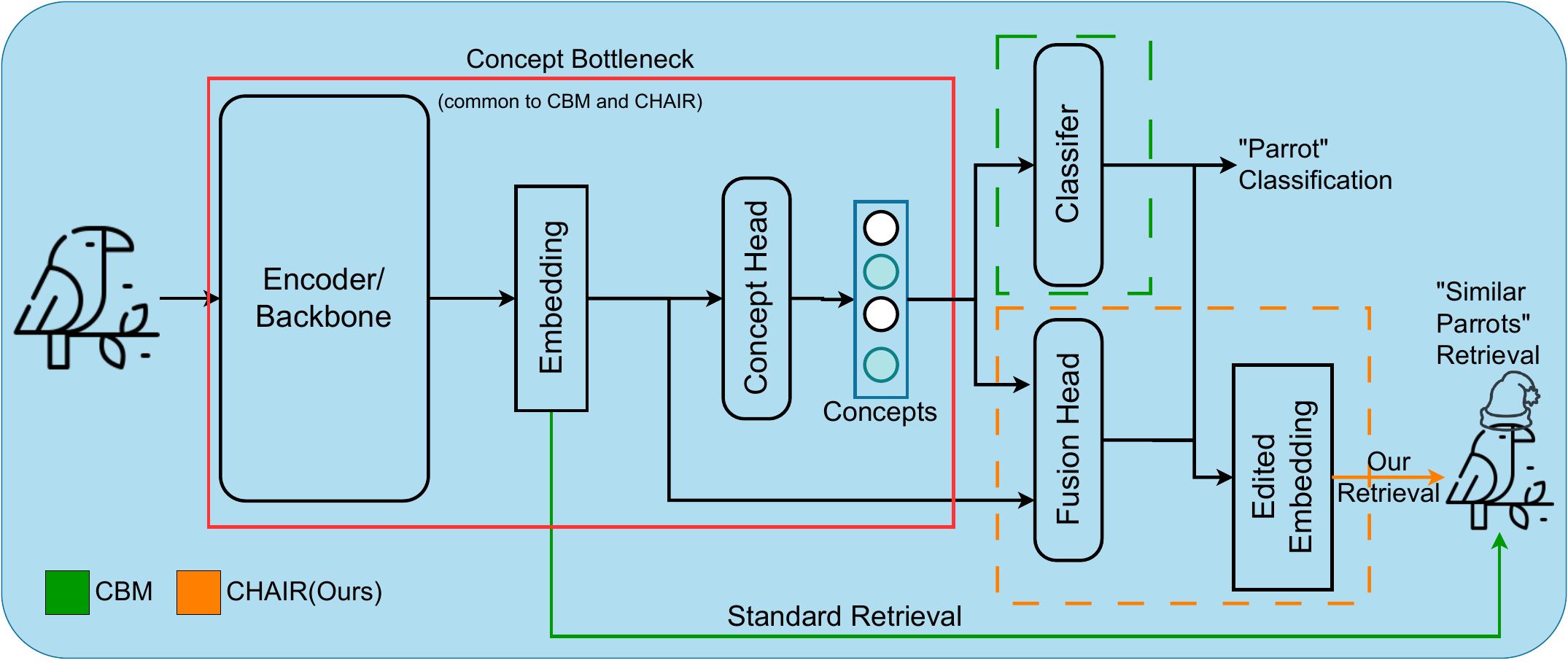}
    \caption{High-level overview of FusionCBM. Our proposed CBM architecture enables \textit{editing} the embeddings using concepts to enable learning better representations and thus improving image retrieval.}
    \label{fig:fusion}
\end{figure*}

We have established the importance of enabling human-AI collaboration for tasks beyond classification. Keeping in mind \textbf{(RQ2)}, which asks \textit{``How can we augment CBMs to enable human interventions on the representations"} and \textbf{(RQ3)} that asks \textit{``How can we enable different levels of expertise for intervention"}, %
we now present \methodname, a two-stage CBM-like architecture that helps address these questions. This section is organized into two parts: \textbf{(a) Architecture} and \textbf{(b) Training}. \\

\paragraph{Architecture:} Figure \ref{fig:fusion} illustrates our proposed \methodname architecture. Let $\zeta$ indicate the encoder that generates embeddings $\mathbf{z}$ from the input image $\mathbf{x}$. $\phi(x)$ denotes the concept head that generates the concept vector $\mathbf{c}$ from the input image $\mathbf{x}$ and $\psi(c)$ denotes the classification head that outputs the final class label $\mathbf{y}$. The \textit{Concept Bottleneck} comprises of the encoder and the concept head, that is $\phi(\zeta(\mathbf{x}))$. The final class prediction in a CBM is as follows:

\begin{equation}
    \mathbf{y_{CBM}} = \psi(\phi(\zeta(\mathbf{x})))
\end{equation}

We propose a simple two-stage modification to the vanilla CBM architecture that enables \textit{integrating} human interventions to improve retrieval. Specifically, as illustrated in Figure \ref{fig:fusion}, our architecture introduces a \textbf{Fusion Head}, whose output is an additional edited embedding alongside the predicted class. The \textbf{Fusion Head} comprises a concept-to-embedding projection layer - a linear layer ($\omega$) that projects the predicted concepts into the same dimensional space as the embedding ($z$), and a classifier. To incorporate the concepts into the embedding, we add the projection and the previous embedding to give us an \textit{edited} embedding, similar in spirit to a residual connection \cite{he2016deep}, thus the name Fusion Head. The objective of this \textit{fusion} is to be able to learn a \textit{meaningful} and \textit{better} representation when these concepts are \textit{corrected}, which helps us address \textbf{(RQ2)}, that is, enabling concept encoding onto the generated embeddings. Lastly, the classifier ($\psi$) utilizes this edited embedding for the final classification, thus preserving the same ability as CBMs to incorporate interventions for improving classification. \\

\paragraph{Training:} Given we have an architecture capable of incorporating human intervention in the latent space for image retrieval and classification, we now introduce a two-stage training that learns better embeddings with intervention while maintaining classification accuracy. Note that we adopt the standard classification loss function for image retrieval as used in \cite{sharif2014cnn}, which is shown to extend to tasks beyond classification. Furthermore, this helps us understand how our architecture performs against similar models trained on the same loss functions. %

\textbf{Stage 1:} From \cite{sharif2014cnn}, we know that the embedding $z$ obtained from training on \texttt{cross-entropy} like loss functions provides a good baseline embedding to enable image retrieval. Leveraging the fusion head architecture, the learned \textit{edited} embedding encodes information from the already performant embeddings ($z$) and concepts. Mathematically, the fusion head consists of the $\phi(z)$ to generate concepts $\mathbf{c}$. $\mathbf{c}$ is then projected to the same space as $\mathbf(z)$ using a simple linear layer $\omega(c)$. Consequently, this projection is added to the original embedding $\mathbf(z)$ to generate the final edited embedding $\mathbf{z'}$. Our proposed fusion head then utilizes a different classification head $\psi'(z')$ to predict the final class label $\mathbf{y'}$ as follows:

\begin{equation}
    \mathbf{y_{CHAIR}} = \psi'(\zeta(\mathbf{x}) + \omega(\phi(\zeta(\mathbf{x}))))
\end{equation}

The goal of Stage 1 is to train the concept-to-embedding projection layer $\omega$ to enable learning better \textit{edited} embeddings $z'$, which improves classification and image retrieval.

\textbf{Stage 2:} Having now trained the concept-to-embedding projection layer, our model is capable of incorporating corrections in the latent space. Referring back to \textbf{(RQ3)}, which aims to include different levels of expertise, we posit that Stage 1 training alone is insufficient to achieve this ability. Hence, we introduce this stage that performs random interventions on a select number of concepts for each mini-batch during training. This random correction of concepts simulates varying levels of expertise while training the \textit{edited} embedding to learn better representations under partial interventions. Note that the concept head here is frozen to preserve the activation values that signify \textit{presence} or \textit{absence} of a concept, similar to \cite{koh2020concept}.

\textbf{Intuition: } Revisiting the requirements of a good human-AI collaborative model, the reasoning behind our contribution is three-fold: (a) our architecture helps us retain the same abilities of CBMs to perform classification, (b) our Stage 1 training allows training $\omega$, the concept-to-embedding projection layer, to enable incorporating concepts into the edited embedding (addressing \textbf{(RQ2)} and (c) Stage 2 training allows for learning quality embeddings under variable intervention, thus addressing \textbf{(RQ3)}. 

We adopt the two training modes outlined in \cite{koh2020concept}: \texttt{Sequential} (shortened as \texttt{Seq} hereafter), where the concept and classification heads are trained separately, and \texttt{Joint} where the heads are trained all at once. Note that these training modes mainly differ in \texttt{Stage 1} training in the loss function. Algorithms \ref{alg:training} and \ref{alg:functions} outline training our proposed \methodname model for both  modes. More specifically, $class\_loss$ and $concept\_loss$ correspond to the same loss functions utilized in the original CBM training. While we observe the same performance pattern for the \texttt{sigmoid} activation function as noted by \cite{koh2020concept}, all of the results we report here are from using the \texttt{ReLU} activation and \texttt{cross-entropy} loss for $class\_loss$ and individually for each $concept\_loss$. Furthermore, the \texttt{intervention\_values} function enables calculating the activation values for concepts that indicate their \textit{presence} or \textit{absence}. We utilize the training data to calculate these values, similar to \cite{koh2020concept}.

\textbf{Intervention: } Intervention in Stage 2 is performed by sampling from a uniform distribution, since we do not assume any level or category of expertise from the humans  in the human-AI team. Therefore, all evaluations with interventions also assume \textit{expertise} on a percentage of the available concepts  sampled uniformly.

\begin{algorithm}[t]
\caption{Training \methodname model}
\label{alg:training}
\kwInput{Training data: $\mathbf{D}=[(x,c,y)]$, mode} 
\kwInitialize{$\phi,\omega,\psi$, $\zeta=$pre-trained,$p_{int}$}
\StageOne{
    \For{$(x_i,c_i,y_i)$ in $\mathbf{D}$}{
        $z_i \gets \zeta(x_i)$  \algorithmiccomment{Get Embedding}
        $c'_i \gets \phi(z_i)$ \algorithmiccomment{Get Concepts}
        $loss = concept\_loss(c_i, c'_i)$
        \If{mode == "sequential"} {
            $z'_i = z_i + \omega(c'_i)$ \algorithmiccomment{Get edited embedding}
            $y'_i = \psi(z'_i)$
            $loss \mathrel{+}= class\_loss(y_i, y'i)$
        }
        \texttt{loss.backward()}
    }
}
\StageTwo{
    \kwFreeze{$\zeta,\phi$}
    \If{mode == "sequential"} {
        \texttt{reset\_weights}($\omega$)
    }
    $c_{int} \gets \texttt{intervention\_values}(\zeta,\phi,D)$
    \For{$(x_i,c_i,y_i)$ in $\mathbf{D}$}{
        $p_i \gets $ \texttt{torch.rand(1)} \algorithmiccomment{Partial intervention}
        $\hat{c}_i \gets$ \texttt{concept\_intervention}($z_i$,$\phi$,$c_i$,$p_i$,$c_{int}$)
        $z"_i \gets \omega(\hat{c}_i) + z_i$
        $y"_i \gets \psi(z"_i)$
        $loss = class\_loss(y_i,y"i)$
        \texttt{loss.backward()}
    }
}
\Return{$\zeta,\phi,\omega,\psi$}
\end{algorithm}

\begin{algorithm}
\SetAlgoLined
\caption{Stage 2 Intervention functions}
\label{alg:functions}
\SetKwFunction{FMain}{\texttt{intervention\_values}}
\SetKwProg{Fn}{Function}{:}{}
\Fn{\FMain{$\zeta,\phi,D$}}{
    \kwInitialize{$c_{int}^{max} = [], c_{int}^{min} = []$}
    $x,c,y = D$ \algorithmiccomment{Training data}\\
    $z \gets \zeta(x)$ \\
    $c' \gets \phi(z)$ \\
    \For{$c'_i$ in $c'$}{ 
        \algorithmiccomment{activations when concept is present and absent} \\
        $c_{int_i}^{max} \gets \texttt{top-95-percentile}(c'_i)$ \\
        $c_{int_i}^{min} \gets \texttt{bottom-5-percentile}(c'_i)$ 
    }
    $c_{int} = [c_{int}^{max}, c_{int}^{min}]$
    \Return{$c_{int}$}
}

\SetKwFunction{FMainTwo}{\texttt{concept\_intervention}}
\SetKwProg{Fn}{Function}{:}{}
\Fn{\FMainTwo{$z$, $\phi$, $c$, $p$, $c_{int}$}}{
    $c' \gets \phi(z)$ \\
    $idx = \texttt{torch.randperm}(c)[:p*\text{len}(c)]$ \\
    $c_{max} = c_{int}[0][idx]$  \algorithmiccomment{Simulates partial correction}\\ 
    $c_{min} = c_{int}[1][idx]$ \\
    $c'[idx] = \texttt{torch.where}(c[idx] == 1, c_{max}, c')$ \\
    $c'[idx] = \texttt{torch.where}(c[idx] == 0, c_{min}, c')$ \\
    $\hat{c} \gets c'$ \\
    \Return{$\hat{c}$}
}
\end{algorithm}

\section{Results}
\label{sec:results}
\subsection{Datasets and Evaluation}
We conduct experiments on two real-world datasets (similar to \cite{EspinosaZarlenga2022cem}), the Caltech-UCSD-Birds-200-2011 (CUB) dataset \cite{wah_branson_welinder_perona_belongie_2011} and Large-scale CelebFaces Attributes dataset (CelebA) \cite{liu2015faceattributes} to demonstrate the effectiveness of our proposed architecture. Specifically, we utilize the CUB dataset to measure performance in both classification and image retrieval tasks, while the CelebA dataset is used for classification only. The CUB dataset comprises $\mathbf{n}=11,788$ bird images belonging to a total of 200 possible species. There are 112 binary concepts associated with each image. We follow the same experimental setup for the classification task as detailed in \cite{koh2020concept}, where the dataset is divided into training, validation, and testing. Furthermore, we follow the data split established in the literature for image retrieval. The training data here consists of bird images from the first 100 classes. The images from the next 100 unseen classes are then used to create the \textit{gallery}, which is used to retrieve images and measure performance. In contrast, each image in the CelebA dataset consists of 40 concept labels, and we utilize 1000 classes for this classification task. We note that the CelebA dataset was not used in evaluating retrieval, as previous works measure retrieval based on binary concepts, which deviates from the main target application of our work. We use ResNet-18 \cite{he2016deep} pre-trained on ImageNet \cite{deng2009imagenet} for all experiments and train on a single \texttt{NVIDIA Tesla V100 16GB} GPU. \footnote{Code (written in Python using PyTorch \cite{paszke2019pytorch}) and instructions to reproduce these results can be found here: \href{https://github.com/realize-lab/CHAIR}{https://github.com/realize-lab/CHAIR}}

\subsection{Retrieval Performance}
\begin{figure}[ht!]
    \centering
    \includegraphics[width=\linewidth]{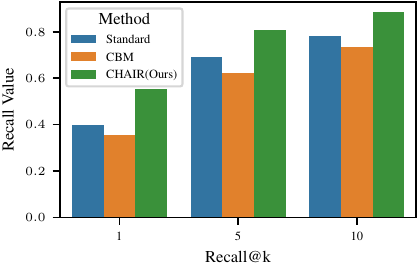}
    \caption{Comparison of the baseline retrieval performance (without any intervention, if possible) of the standard ResNet model, vanilla CBM, and the proposed \methodname model.}
    \label{fig:retrieval-results}
\end{figure}

\begin{figure}[ht!]
    \centering
    \includegraphics[width=\linewidth]{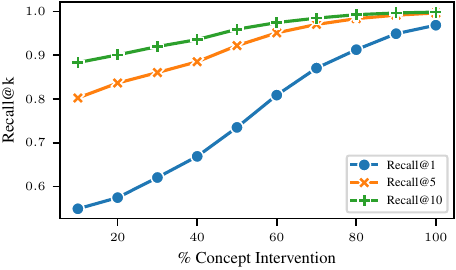}
    \caption{\textbf{Intervention vs \texttt{Recall@k} for \methodname} \textit{Accurate} interventions on an increasing subset of concepts helps extract better quality representations, thus improving retrieval performance.}
    \label{fig:retrieval-vs-intervention}
\end{figure}

Figure \ref{fig:retrieval-results} shows how \methodname performs against a standard ResNet-18 model and CBM. A typical CBM with the same underlying ResNet-18 architecture performs slightly worse than the standard ResNet model, which aligns with the behavior noted by \cite{koh2020concept} in classification. Our model provides 15-20 \% $Recall@k$ improvement for $k=[1,5,10]$, indicating that our model has learned better-unedited embeddings as opposed to the embeddings from the standard and CBM models.  Furthermore, we randomly select $p\%$ of concepts for every image in the \textit{gallery} and each image during \textit{query} to investigate how intervention improves retrieval and evaluate the $Recall@k$ performance, as shown in Figure \ref{fig:retrieval-vs-intervention}. We observe that as the number of interventions (corrections) increases, the retrieval performance improves, eventually reaching near-perfect $Recall@k$. A key observation that emerges from this plot is that when all the predicted concepts are set to their correct values, the method achieves nearly perfect recall when $k=10$.

\begin{figure*}[h]
    \centering
    \includegraphics[width=\linewidth]{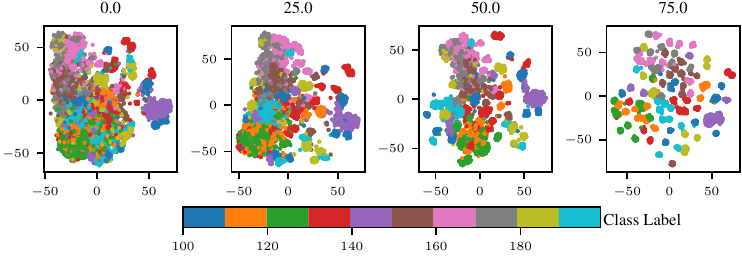}
    \caption{t-SNE visualization of the embeddings generated at increasing levels of intervention (\%level indicated at the top of each sub-plot)}
    \label{fig:tsne}
\end{figure*}

\begin{figure}[H]
    \centering
    \includegraphics[width=\linewidth]{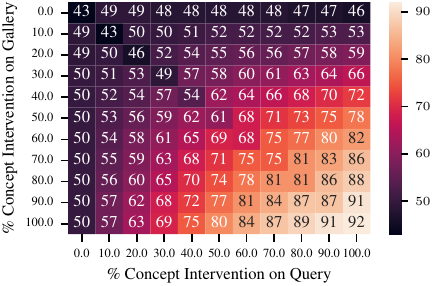}
    \caption{Heatmap of $RecallAccuracy@10$ (\%) for variable levels of intervention in the gallery and query time}
    \label{fig:labels-at-top-10}
\end{figure}

In the real world, it is possible that the labelers constructing the gallery embeddings could be less experienced than the users  %
during query time (especially when the gallery is crowdsourced). Hence, we demonstrate how different levels of intervention in the query and gallery images impact the accuracy of our proposed retrieval architecture. Figure \ref{fig:labels-at-top-10} shows a heatmap of \% interventions on query images and gallery images on the x and y axes, respectively, and the \texttt{RecallAccuracy@10} value for each intervention pair. We observe that even when the gallery is constructed with no intervention, any amount of human intervention during query time helps improve retrieval performance. Notably, this trend emerges across all interventions during query time when the interventions of the gallery embeddings are kept fixed.  Note that we do not plot a similar heatmap for \texttt{Recall@k}, since our method already performs %
well without any intervention. 

\subsection{Importance of Stage 2}
\begin{figure}[H]
    \centering
    \includegraphics[width=0.9\linewidth]{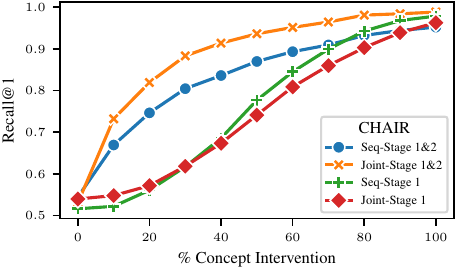}
    \caption{Stage 2 training allows better performance when the intervention is incomplete.}
    \label{fig:stage-two-imp}
\end{figure}

Recall that Stage 2 was introduced to help improve performance under partial intervention. To measure the impact of Stage 2, we calculate the recall values of both the training modes (\texttt{Seq} and \texttt{Joint}, as defined in Section \ref{sec:fusion}) perform with and without Stage-2 training for each subset of intervention in Figure \ref{fig:stage-two-imp}. This figure clearly shows the benefit acquired when Stage 2 training is employed, especially in the 0-80\% range, where we obtain 5\%-25\% improvement in recall depending on the intervention. Note that the \texttt{Seq.} performance here without Stage 2 implies the classification head is trained on predicted concept activation values without any intervention. 

\subsection{Quality of Edited Representations}

In order to visualize the representations at each intervention level, we use t-Distributed Stochastic Neighbor Embedding (t-SNE), a dimensionality reduction technique that helps visualize high-dimensional data, which in our case are the embeddings that are used for retrieval. We extract embeddings for each image present in our test data (all the classes in the test data are unseen during training) and apply t-SNE to reduce them to 2 dimensions. We then plot these reduced embeddings, as shown in Figure \ref{fig:tsne} for different levels of intervention. With accurate intervention on increasing random subsets of the concepts, the clusters for each class become more distinct. These visual representations also help provide evidence to support how the recall performance under no intervention is much lower when compared to performance under intervention.

\subsection{Classification Performance}
\begin{figure}[H]
    \centering
    \includegraphics[width=0.9\linewidth]{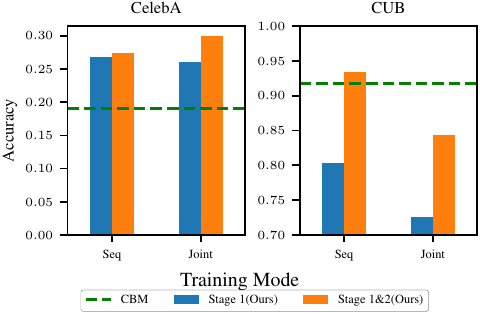}
    \caption{Comparing our proposed model with different stages of training against CBM in classification tasks with maximum intervention. Our model consistently outperforms CBM in classification.}
    \label{fig:classification-results}
\end{figure}
As referred to in \textbf{(RQ2)}, our model must be able to match the performance of CBMs in classification and allow intervenable retrieval so that practitioners do not have a tradeoff when selecting the architectures. Figure \ref{fig:classification-results} compares the classification performance of both sequential (\textit{Seq}) and \textit{joint} training modes under perfect concept intervention against the best-performing classifying CBM on both the CUB and the CelebA datasets. It is evident that our proposed \methodname model outperforms the standard best-performing CBM in these real-world datasets. Furthermore, this plot shows how Stage 2 helps improves classification performance, thus showing its importance beyond improving retrieval under partial intervention.

\section{Conclusion and Future Work}

In this work, we first establish that CBMs, a model that allows for human-AI collaboration, underperform when compared to standard neural networks on retrieval tasks. We propose \methodname, a modification of both CBM architecture and training strategy, which can a) incorporate human input in the form of concept correction for image retrieval, b) allow varying levels of human input and expertise, and c) significantly improve retrieval performance while maintaining similar classification performance to vanilla CBMs. Furthermore, we show that the quality of the embeddings generated with some corrections performs better than the alternatives through t-SNE plots and improved retrieval performance. Finally, it is evident that while both the Seq and Joint training modes perform equally well with a high level of intervention, we recommend choosing the joint training mode for image retrieval-related tasks. Our work enables the expansion of CBMs to domains that move beyond classification, further improving human-AI collaboration in impactful applications such as wildlife population monitoring. 

We envision future work in this area along multiple frontiers: a) incorporating probabilistic concepts, label prediction, and embeddings to capture the uncertainty of the prediction process better \cite{kim2023probabilistic,li2021learning}, b) learning when and how to defer to humans that are interfacing with CBMs to achieve the best possible complementary performance \cite{bondi2022role,mozannar2020consistent}, and c) conducting human studies to understand effective strategies of presenting CBM information to achieve complementarity. 

\section*{Ethical Statement}

Our contributions and future work involve including and complementing human efforts. Hence, we strongly advocate rigorously testing with all stakeholders before deploying \pname-like models.

\section*{Acknowledgements}

We want to thank the reviewers for their insightful comments and suggestions. We also thank the e-Health and Artificial Intelligence (e-HAIL) initiative and the Advanced Research Computing (ARC) at the University of Michigan, Ann Arbor, for their support and for providing computing resources and services, respectively.

\bibliographystyle{named}
\bibliography{ijcai24}

\end{document}